\pdfoutput=1
\documentclass[conference]{IEEEtran}
\usepackage{epsfig}
\usepackage{amsmath}
\usepackage{pdflscape}
\usepackage{graphicx}
\ifCLASSINFOpdf

\else

\fi

\begin{document}

\title{Detecting Moving Objects Using a Novel
Optical-Flow-Based Range-Independent Invariant
}

\author{\IEEEauthorblockN{Daniel Raviv}
\IEEEauthorblockA{College of Engineering and\\ Computer Science\\
Florida Atlantic University\\
Boca Raton, Florida 33431\\
Email: ravivd@fau.edu}
\and
\IEEEauthorblockN{Juan D. Yepes}
\IEEEauthorblockA{College of Engineering and\\ Computer Science\\
	Florida Atlantic University\\
	Boca Raton, Florida 33431\\
	Email: jyepes@fau.edu}
 \and
 \IEEEauthorblockN{Ayush Gowda}
\IEEEauthorblockA{College of Engineering and\\ Computer Science\\
	Florida Atlantic University\\
	Boca Raton, Florida 33431\\
	Email: agowda2019@fau.edu}}

\maketitle

\begin{abstract}
This paper focuses on a novel approach for detecting moving objects during camera motion. We present an optical-flow-based transformation that yields a consistent 2D invariant image output regardless of time instants, range of points in 3D, and the camera’s speed. In other words, this transformation generates a lookup image that remains invariant despite the changing projection of the 3D scene and camera motion. In the new domain, projections of 3D points that deviate from the values of the predefined lookup image can be clearly identified as moving relative to the stationary 3D environment, making them seamlessly detectable.

The method does not require prior knowledge of the camera’s direction of motion or speed, nor does it necessitate 3D point range information. It is well-suited for real-time parallel processing, rendering it highly practical for implementation. We have validated the effectiveness of the new domain through simulations and experiments, demonstrating its robustness in scenarios involving rectilinear camera motion, both in simulations and with real-world data. This approach introduces new ways for moving objects detection during camera motion, and also lays the foundation for future research in the context of moving object detection during six-degrees-of-freedom camera motion.

\end{abstract}

\IEEEpeerreviewmaketitle

\section{Introduction}
A fundamental challenges in computer vision is the detection of moving objects in a 3D environment while the observer is in motion \cite{cutting1995we}.

The problem is particularly challenging because both the projection of the 3D environment and the moving objects onto the images of the moving camera constantly generate changing optical flow. Consequently, distinguishing moving objects from the 3D surroundings becomes a formidable task \cite{dupin2013motion}.

The complexities further intensify when the objective is to identify multiple moving objects within a dynamic environment \cite{wang2007simultaneous}. Acknowledging this challenge, this paper introduces a transformative approach that can potentially reshape the landscape of moving object detection \cite{azim2012detection}.

This paper presents a novel method for detecting moving objects during camera motion. It creates a consistent 2D invariant image of a stationary environment using functions of optical flow components \cite{horn1981determining}, making moving objects easily identifiable irrespective of the stationary environment’s 3D structure \cite{ozyecsil2017survey}. It works without prior knowledge of camera speed, and also when the camera accelerates or decelerates.  The transformation is pixel-based and parallel in nature, making it suitable for real-time processing. Validation through simulations and experiments demonstrates its robustness during camera rectilinear motion, which has the potential to transform object detection \cite{arnold2019survey} and open different avenues for six-degrees-of-freedom camera motion research.
This research could potentially open the door to a new era of visual motion invariant-based solutions for some difficult problems.

\section{Method}
In this paper, we assume rectilinear motion of the camera with unknown and potentially changing speed. To simplify the explanation of the core idea behind the approach, we discuss the case where the optical axis of the camera coincides with its direction of motion. 

\begin{figure}
	\centering
	{\epsfig{file = 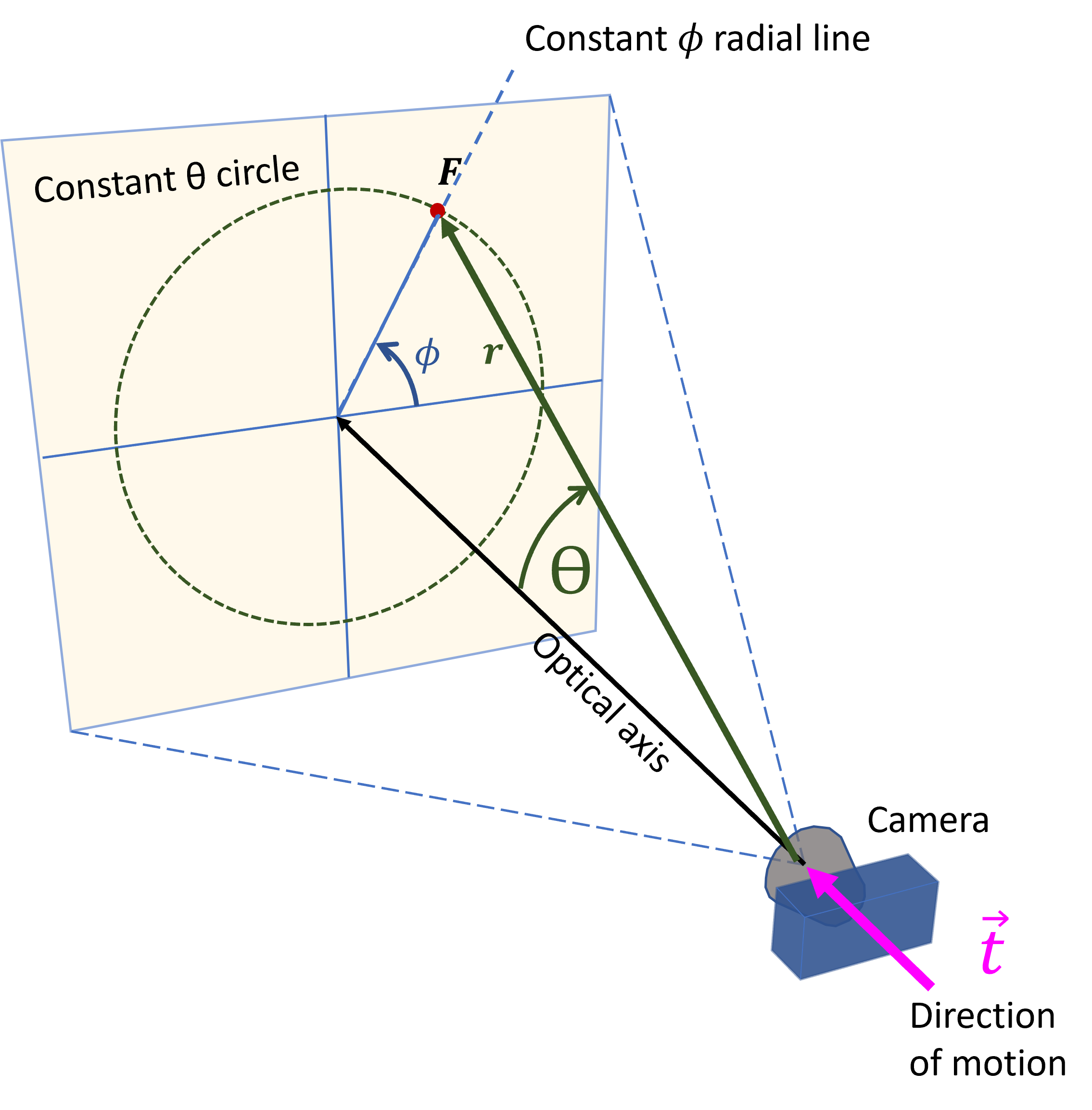, width = 8cm}}
	\caption{Coordinate System.}
	\label{Fig:coordinateSystem}
\end{figure}

Refer to Figure \ref{Fig:coordinateSystem}. Each point in 3D can be described by its range and two angles: $\theta$ and $\phi$. When the camera translates, the projections of points onto the image either move radially away from the Focus of Expansion (FOE) or toward the Focus of Contraction (FOC) \cite{jain1983direct}.\\

\subsection{Main ideas}
\subsubsection{Range Independent Invariant}
The $\dot{\theta}$ optical flow can be decomposed into vertical and horizontal components, $\dot{\theta}\sin{\phi}$ and $\dot{\theta}\cos{\phi}$, correspondingly. The ratio of these two optical flow values at a given ($\theta$, $\phi$) set of angles is unitless, can be a-priori computed, and always have the \emph{same value independent of the range of the point}. This allows us to construct a lookup image \emph{consistent across all 3D environments}, regardless of the camera’s speed. In other words, for any rectilinear motion of the camera in any stationary 3D environment, the result of the transformation is a lookup image, ensuring invariance in both range and speed.\\

\subsubsection{Identifying moving objects}
In the newly established lookup image invariant domain, the projections of 3D moving points exhibit varying ratios of optical flow values. These ratios deviate from the values in the predefined lookup image and can be \emph{easily identified as objects} in motion relative to the stationary 3D environment, making them seamlessly detectable \cite{azim2012detection}.

\section{Results}
This section deals with two different types of results, Unity based-simulation and real data as obtained from a moving robot.

\subsection{Simulation}
Using Unity \cite{borkman2021unity} as a simulation platform, the transformation was tested under various conditions, including different 3D environments and a moving camera. The results consistently demonstrated the transformation's ability to provide a constant lookup image, independent of the environment's complexity. 

\begin{figure}
	\centering
	{\epsfig{file = 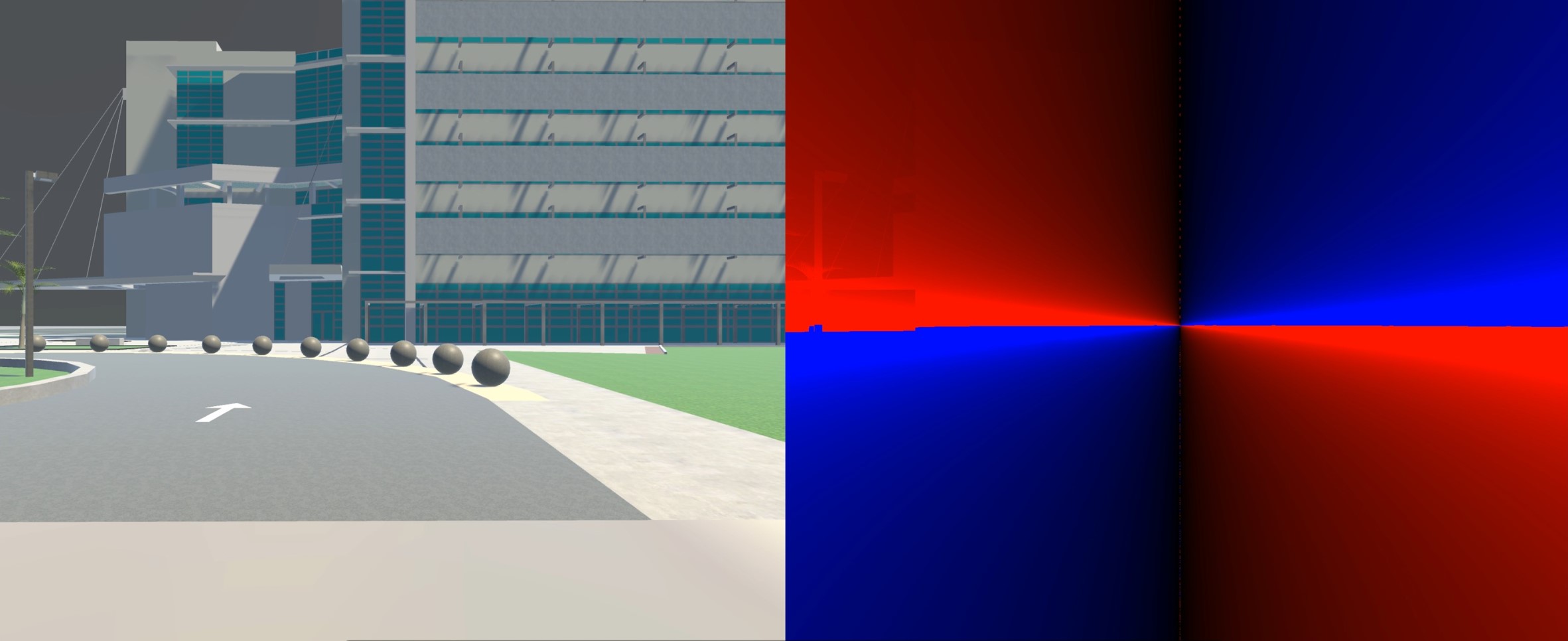, width = 8.5cm}}
	\caption{Simulation with Stationary Environment.}
	\label{unity1}
\end{figure}

Refer to Figure \ref{unity1}. The left image shows a snapshot obtained from a moving video during camera motion in a 3D stationary environment. The right image is the color-coded representation of the result of the ratio of the optical flow components, which, as explained earlier, is the lookup image. The brightness of the colors represents the value as obtained from the transformation. Red and blue colors correspond to different signs of the transformation ratio.

Note that the intersection of the four quadrants in the right part of Figure \ref{unity1} indicates the location of the focus of expansion (FoE).

\begin{figure}
	\centering
	{\epsfig{file = 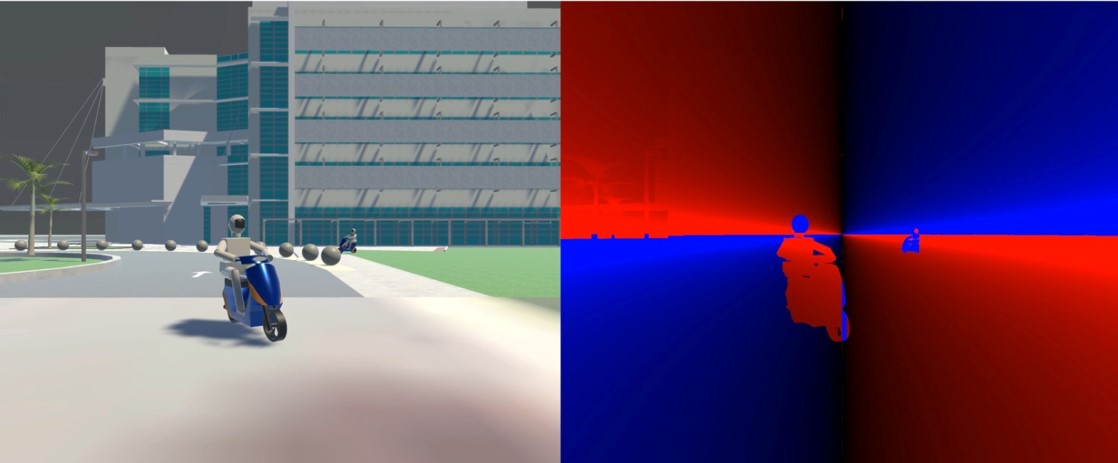, width = 8.5cm}}
	\caption{Simulation with Moving Objects.}
	\label{unity2}
\end{figure}

When introduced to moving objects within an otherwise stationary environment, the method successfully detected the object's presence. The disruption in the optical flow ratio caused by the moving objects is clearly visible as can be seen in Figure \ref{unity2}. 

The sole distinction between the scenarios in Figure \ref{unity1} and Figure \ref{unity2} is the addition of moving objects in the latter. Additionally, all points in the right section of Figure \ref{unity2}, unrelated to these moving objects, match the color of the corresponding points in Figure \ref{unity1}, the lookup image.

\subsection{Real Data}

\begin{figure}
	\centering
	{\epsfig{file = 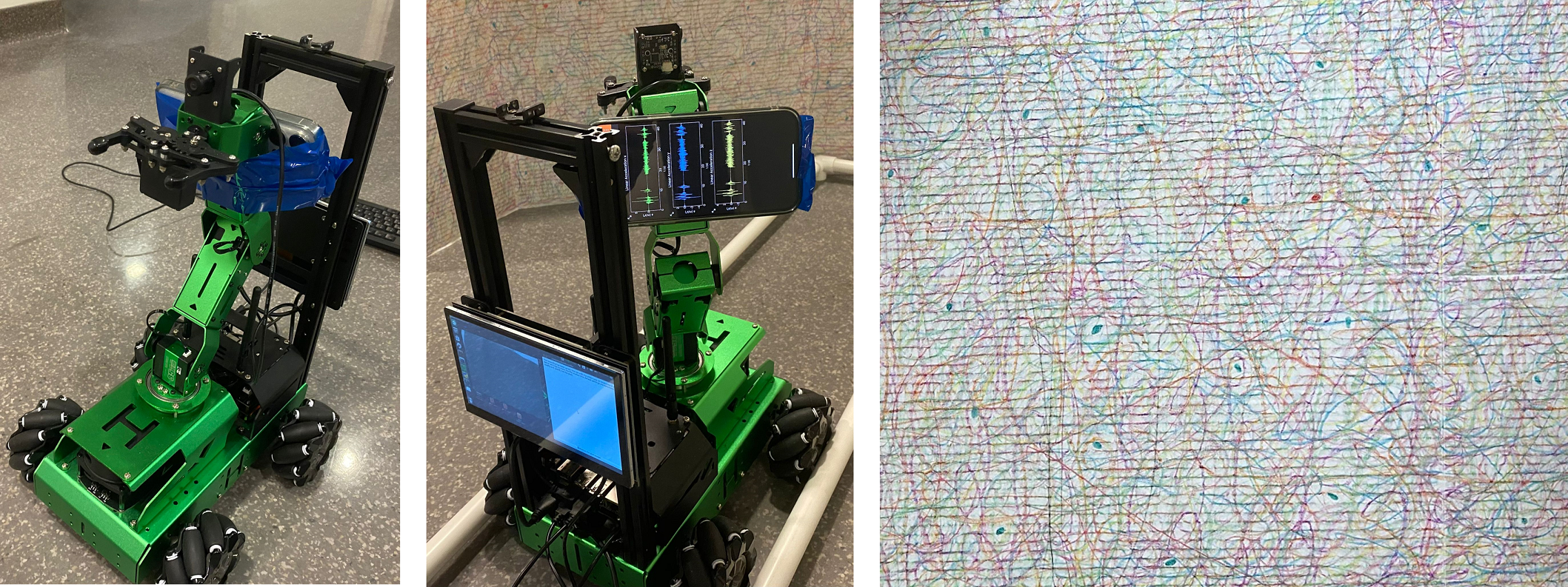, width = 8.5cm}}
	\caption{The JetAutoPro robot platform is shown along with the camera on the left. The PVC track is shown in the center along with the back view of the robot. The tri-Fold board with a static pattern is shown on the right. }
	\label{robot}
\end{figure}

For real-world testing, the JetAutoPro robot by HiWonder Shenzen was employed. The robot’s arm-mounted camera was configured to be approximately parallel to the floor, and perpendicular to a tri-Fold board featuring a static pattern. In order to maintain rectilinear motion, a simple PVC track was created, as seen in Figure \ref{robot}.
Movement commands were executed using adapted built-in modules to ensure constant speed and for capturing the video. The frames were processed in real-time with an onboard NVIDIA Jetson Nano using optical flow obtained by the Farneback algorithm \cite{farneback2003two}. 

\begin{figure}
	\centering
	{\epsfig{file = 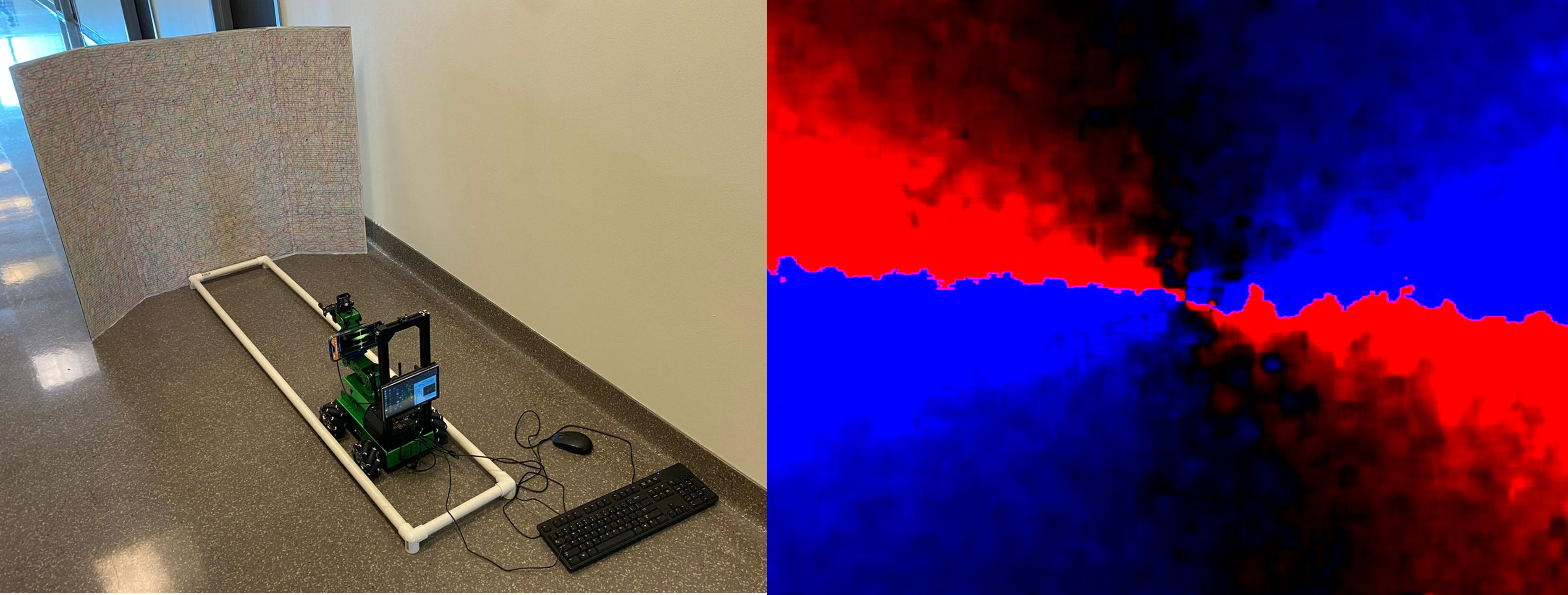, width = 8.5cm}}
	\caption{On the left, the robot is moving perpendicular to the tri-fold board. On the right, the transformation from this motion is shown.}
	\label{realStraight}
\end{figure}

Real-world testing involved two distinct movement paths, both at a constant speed. In the first scenario (see left part of Figure \ref{realStraight}), the robot moved perpendicular to the tri-fold board, with the camera's optical axis also perpendicular. The result of the transformation can be observed in the right section of Figure \ref{realStraight}.

\begin{figure}
	\centering
	{\epsfig{file = 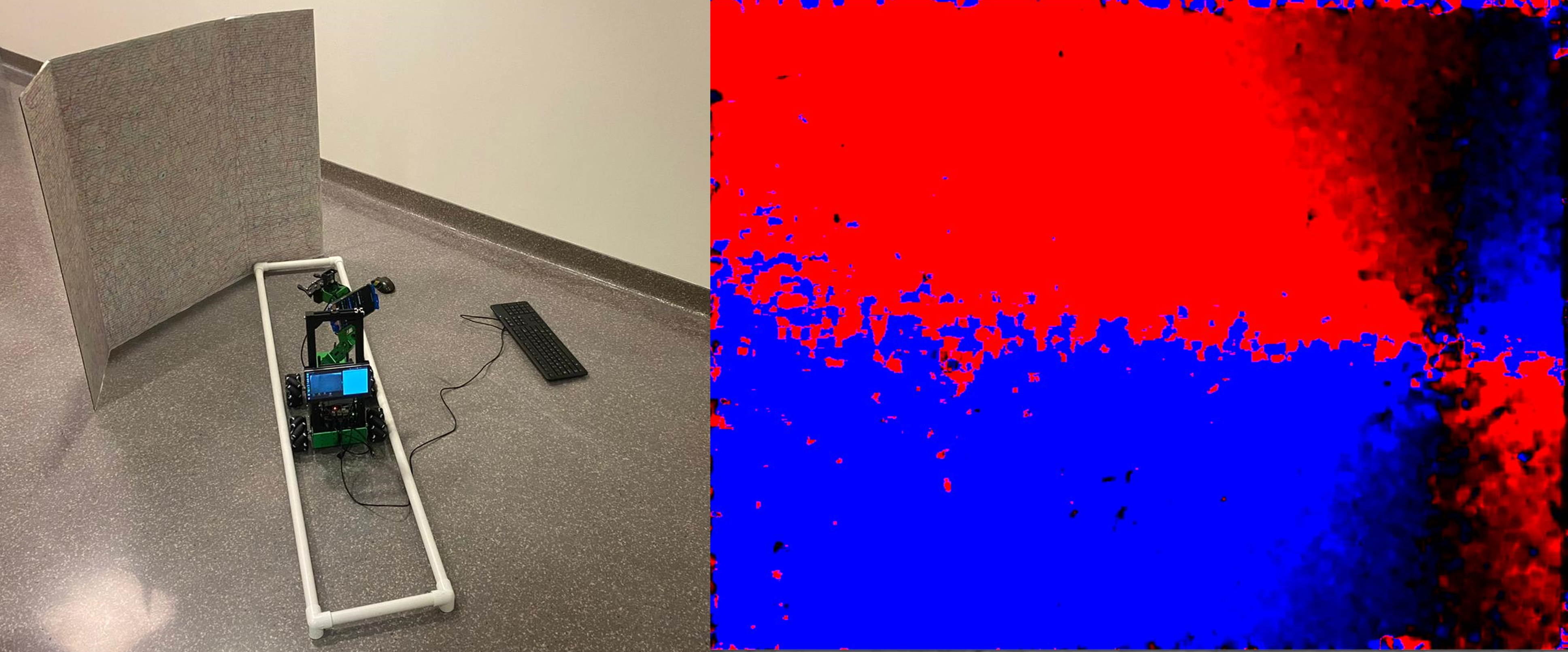, width = 8.5cm}}
	\caption{On the left, the robot is moving slanted to the tri-fold board. On the right, the transformation from this motion is shown.}
	\label{realTilt}
\end{figure}

In the second scenario (see left part of Figure \ref{realTilt}), the robot moved in a slanted angle relative to the tri-fold board, however the camera’s optical axis remained perpendicular to the board. The result of the transformation due to this motion is shown in the right section of Figure \ref{realTilt}.

By isolating the horizontal and vertical components of the estimated optical flow for the Farneback algorithm, we obtained the result of the transformation that looks very similar to the lookup image. While other optical flow methods such as RAFT \cite{teed2020raft} are more accurate than the Farneback, the latter was used due to its real time advantages and low computational requirements. 

 When the robot moves diagonally it causes a shift in the location of the focus of expansion, as can be seen in the right image of Figure \ref{realTilt}. While the reference lookup image remains unchanged in content (at least theoretically), its position appears shifted to the right due to the angular differences between the camera optical axis and the direction of motion. 
 
\section{Conclusion and Future Work}
In this paper, we have presented a novel invariant domain that can be employed for the detection of moving objects during camera rectilinear motion. This domain yields identical lookup image for any 3D stationary environment. This reference image can be subtracted from the actual data as obtained in the new domain, theoretically isolating only the influence of the moving objects, and thereby enabling the detection of the moving objects.
Using simulation results, we have shown that moving objects can be easily detected. This approach requires one camera only, is pixel based and suitable for parallel processing. 
The results obtained from a real video are highly encouraging and demonstrate the potential of the method for future applications.

While our transformation shows promise, it is not without limitations. One such limitation is the rectilinear motion of the camera. We are working on extending the method to include any six-degrees-of-freedom motion of the camera. Filtering and smoothing and processing the real data in the new domain over time can increase the accuracy and the utility of this method.

\section*{Acknowledgment}
This work was supported in part at the Technion through a fellowship from the Lady Davis Foundation. The authors would like to thank Michael Levine for his continued support of this project. We also thank Clint Hatcher for proofreading the paper and suggesting very useful comments.


%

\bibliographystyle{IEEEtran}
\bibliography{IEEEabrv,timeMapping}


\end{document}